\documentclass[10pt,twocolumn,letterpaper]{article}

\usepackage{cvpr}              
\usepackage{multirow}
\usepackage{tabularx}
\usepackage{natbib}

\newcolumntype{C}{>{\centering\arraybackslash}X}
\newcolumntype{L}{>{\raggedright\arraybackslash}X}
\newcolumntype{R}{>{\raggedleft\arraybackslash}X}

\definecolor{cvprblue}{rgb}{0.21,0.49,0.74}
\usepackage[pagebackref,breaklinks,colorlinks,allcolors=cvprblue]{hyperref}

\def\paperID{} 
\def\confName{}
\def\confYear{2025}

\title{Understanding and Improving Training-Free AI-Generated Image Detections with Vision Foundation Models}

\author{Chung-Ting Tsai$^{1,2}$, Ching-Yun Ko$^1$, I-Hsin Chung$^1$, Yu-Chiang Frank Wang$^2$, Pin-Yu Chen$^1$ \\
$^1$IBM Research \quad $^2$National Taiwan University \\
{\tt\small b09901020@ntu.edu.tw}
}

\begin{document}
\maketitle
\begin{abstract}
The rapid advancement of generative models has introduced serious risks, including deepfake techniques for facial synthesis and editing. Traditional approaches rely on training classifiers and enhancing generalizability through various feature extraction techniques. Meanwhile, training-free detection methods address issues like limited data and overfitting by directly leveraging statistical properties from vision foundation models to distinguish between real and fake images. The current leading training-free approach, RIGID, utilizes DINOv2’s sensitivity to perturbations in image space for detecting fake images, with fake image embeddings exhibiting greater sensitivity than those of real images. This observation prompts us to investigate how detection performance varies across model backbones, perturbation types, and datasets. Our experiments reveal that detection performance is closely linked to model robustness, with self-supervised (SSL) models providing more reliable representations. While Gaussian noise effectively detects general objects, it performs worse on facial images, whereas Gaussian blur is more effective due to potential frequency artifacts. To further improve detection, we introduce Contrastive Blur, which enhances performance on facial images, and MINDER (MINimum distance DetEctoR), which addresses noise type bias, balancing performance across domains. Beyond performance gains, our work offers valuable insights for both the generative and detection communities, contributing to a deeper understanding of model robustness property utilized for deepfake detection.
\end{abstract}    

\section{Introduction}
\label{sec:introduction}

\renewcommand{\thefootnote}{\fnsymbol{footnote}}
\footnotetext[1]{This work was done while Chung-Ting Tsai was a visiting
researcher at IBM Thomas J. Watson Research Center.}
\renewcommand{\thefootnote}{\arabic{footnote}}

Recent advancements in generative models, such as Stable Diffusion \cite{sd}, have made creating realistic visuals more accessible, aided by personal GPUs and tools like LoRA \cite{hu2021lora} for fine-tuning. However, as these models produce increasingly lifelike images, new misuse cases arise, such as deepfakes for face-swapping, which raise concerns about misinformation and privacy. Additionally, generative models can be used to create harmful content through malicious prompts, and intellectual property issues emerge when AI-generated content mimics existing works, risking copyright infringement. These concerns highlight the need for research into ethical safeguards for AI-generated content.

Detecting synthetic facial images has been widely studied in response to generative advances. With models guided by text prompts, synthetic images—across both specialized and general domains—are increasingly hard to distinguish from real ones. Many detection models struggle to generalize to unseen generative methods, though some improve generalizability through data augmentation \cite{general1, general2} or by exploiting fake image artifacts, such as frequency patterns \cite{freq1, freq2}.

As generative techniques continue to evolve and advance, producing high-quality, diverse, and comprehensive fake images to train a reliable and generalizable AI-image detector has become more resource-intensive and challenging. This has led to the rise of training-free detection methods, which address the generalization challenge without relying on large datasets, thus avoiding overfitting. While training-free detection methods are often praised for mitigating overfitting, we believe that a significant advantage lies in the insights they provide into underlying mechanisms. For example, the training-free method Aeroblade \cite{ricker2024aeroblade} shows that images generated by latent diffusion models map to a constrained manifold in latent space, while real images are projected to the nearest point, causing increased reconstruction error when passed through an autoencoder.

Another notable work, RIGID \cite{rigid}, proposes a training-free framework for detecting AI-generated images by leveraging sensitivity differences between real and fake images. \citet{rigid} explain that self-supervised models like DINOv2 \cite{dinov2} provide a more holistic perspective, capturing a broader understanding of image content, and thus outperform models like ResNet \cite{resnet} or CLIP \cite{clip}, as shown through Grad-CAM analysis. However, our primary interest lies in understanding why fake images exhibit greater sensitivity. In this work, we further investigate the underlying mechanisms, starting with DINOv2’s ability to produce robust representations. We hypothesize that the vision foundation model captures the distribution of real images, where the model perceives a more stable representation. By applying anomaly detection through noise analysis, the model can effectively distinguish between real and fake images.

In this work, we aim to address the question: \textit{Can the properties identified by RIGID be broadly applied to AI-generated images?} A follow-up question examines how these properties function, or fail to function, in specific cases. We begin by applying RIGID \cite{rigid} to the recently released facial dataset DF40 \cite{df40}, and surprisingly, we discover that Gaussian noise (the recommended operation in RIGID) was ineffective in certain subsets. We then tested alternative noise types and found that blurring was unexpectedly effective. A deeper analysis of the frequency domain revealed that frequency artifacts play a key role in the effectiveness of blurring. Furthermore, we explore other vision foundation models to replace DINOv2 \cite{rigid}, including its predecessors, models of varying scales, and other non-self-supervised learning (SSL) foundation models, reinforcing the idea that the models' robust representations contribute to their detection capabilities.

Building on the frequency artifacts observed in fake facial images, we further develop the method \textit{Contrastive Blur} to increase the distance between perturbed embeddings, thereby enhancing performance. While Gaussian noise is effective for general images and Gaussian blur works well for facial images, we introduce a new detector, \textbf{MINDER} (\textbf{MIN}imum distance \textbf{D}et\textbf{E}cto\textbf{R}), which extends the RIGID framework to achieve more consistent performance across both facial and general image domains.

Our contributions are threefold: 
\begin{itemize}
\item We provide valuable insights through logically structured experiments and analyses, enhancing the understanding of both detection models and images produced by generative models. 
\item We propose \textbf{MINDER}, a novel detector that incorporates minimum distance selection to mitigate detection bias in current perturbation-based detection frameworks. 
\item Our framework extends the existing training-free detection method RIGID \cite{rigid}, achieving the highest overall performance (averaged across facial and general images) among training-free detection methods. 
\end{itemize}
\section{Related Work}
\label{sec:related}

\paragraph{AI-based Image Generation} The evolution of Generative Adversarial Networks (GANs), followed by the rise of diffusion models, has propelled AI-generated content into the public spotlight. StyleGAN \cite{stylegan} marked a major breakthrough, significantly enhancing image quality and providing unprecedented control over generated images. BigGAN \cite{biggan} further improved training stability through regularization techniques, while VQGAN \cite{vqgan} captivated attention with its ability to produce high-resolution outputs. Diffusion models \cite{ddpm} introduced a new paradigm, achieving greater training stability and higher image quality, though with some trade-off in generation time. Numerous follow-up works, such as DDIM \cite{ddim}, DiT \cite{dit}, PixArt \cite{pixart}, ADM \cite{adm}, GLIDE \cite{nichol2021glide}, and VQDM \cite{vqdm}, have expanded on these advancements. These successes have led to widespread commercial applications, with models like the Stable Diffusion series (1–3.5) \cite{sd}, MidJourney versions 1 through 6 \cite{midjourney}, and the recent FLUX.1 \cite{flux}, which have scaled up models and significantly improved image quality.

\paragraph{AI-based Image Detection} AI-generated image detection has long been an important problem. From the GAN era, frequency-based methods have been extensively explored \cite{frank2020leveraging, chandrasegaran2021closer, freq1, freq2}. A significant improvement in generalization was made by \cite{cnneasy}, who showed that a simple detector trained with data from ProGAN \cite{progan} and data augmentation could generalize to images generated by various GAN models. More recently, \cite{corvi2023detection} has tackled the more challenging task of detecting images generated by diffusion models. Recent works like NPR \cite{npr} revisit the up-sampling process used in GANs and variational autoencoders (VAEs) in latent diffusion models, detecting fake images through neighboring pixel relationships. DF40 \cite{df40} trains a CLIP  model \cite{clip} across large and diverse datasets, demonstrating that large-scale data may be a simple but effective method, though costly. AIDE \cite{aide} develops a two-stream framework that leverages both frequency and semantic information.

\paragraph{Training-Free Detection.} Due to the limited generalizability of training-based detection models, an alternative approach known as training-free detection has emerged. This method leverages the capabilities of pre-trained models to capture distinguishing features of both real and fake images. AeroBlade \cite{ricker2024aeroblade} asserts that fake images generated from the latent space of diffusion models naturally exhibit smaller reconstruction errors when passed through autoencoder-based latent diffusion models. RIGID \cite{rigid} capitalizes on the observation that real images are less sensitive to random perturbations in the representation space, using DINOv2 \cite{dinov2} as the vision encoder. Another variation is the weakly-supervised setting in ZED \cite{zed}, a zero-shot detector trained without fake images. It borrows the concept from likelihood-based detection in AI-generated text detection, adapting it within a multi-level super-resolution framework. The promising results of these training-free methods inspire further research into deeper analysis and extensions of this approach.
\section{How Does Model Robustness Facilitate Training-Free Detection?}
\label{sec:understand}

We present a logically structured, step-by-step analysis to understand better the mechanism behind the perturbation-based method, RIGID \cite{rigid}. This includes examining its effectiveness on facial images, exploring various types of noise, performing frequency analysis, and, finally, discussing variations in backbone architectures. The insights gained highlight potential ways to improve current models, as well as the limitations of existing approaches.

\subsection{Robustness of Vision Foundation Model}

Since the Vision Transformer (ViT) \cite{vit} architecture demonstrated outstanding performance and scalability, numerous off-the-shelf vision foundation models, including CLIP \cite{clip}, MAE \cite{mae}, and DINOv2 \cite{dinov2}, have been trained on large-scale datasets and can be further adapted to downstream tasks. While these models are undoubtedly powerful, their robustness in task performance is enhanced through techniques like data augmentation during training. With substantial real-world data, these image encoders learn robust representations of visual inputs, allowing them to remain stable under perturbation. The success of RIGID \cite{rigid} has shown that the robustness of learned real image embeddings can serve as an effective anomaly detection method for fake images. Encoded embeddings of fake images are more sensitive to the perturbed noise than those of real ones, making the cosine similarity between original and perturbed output embeddings a useful metric for identifying AI-generated images. Specifically, the threshold-based detection can be conducted by
\begin{equation}
  y(x) = \textbf{1}\{ \ \mathcal{D} ( f(x), \ f(x + \delta)) \geq \epsilon \ \},
  \label{eq:similarity}
\end{equation}
where $f(\cdot)$ is the representation encoder, $x$ is the visual input, and $\delta$ is the perturbation drawn from a noise or corruption method, such as $\delta_{\text{noise}}$ for Gaussian noise or $\delta_{\text{blur}}$ for the perturbation introduced by the blurring process. $\mathcal{D}(\cdot)$ denotes the distance function, and for simplicity, we may express $\mathcal{D}(f(x), f(x + \delta))$ as $\mathcal{D}(x, \delta)$ in the following sections. The indicator function $\mathbf{1}\{\cdot\}$ assigns a label of 1 to an image classified as fake when the similarity distance exceeds the threshold $\epsilon$. While this method can establish a likelihood ranking between real and fake images, setting the threshold $\epsilon$ requires calibration on a validation set of real images, with a common practice being to fix the false positive rate (FPR) at 5\%.

The authors of RIGID \cite{rigid} have demonstrated that using $\delta \sim \mathcal{N}(0, \sigma^2)$ and $\mathcal{D}(\cdot)$ with cosine distance can achieve significant success—attributing this result to the sensitivity difference between real and fake images—This promising result encourages further investigation into how detection performance varies across different perturbation methods, encoder backbone, and datasets.

\subsection{Gaussian Blur on Deepfake Images}

While RIGID \cite{rigid} has been shown to effectively detect differences between ImageNet \cite{imagenet} images and some AI-generated images, we shift our focus toward a more pressing application scenario: deepfake detection. Unlike ImageNet images, facial images are highly structured, and deepfake generation has been extensively studied, adding complexity to fake facial image detection. In our experiments, we unexpectedly found that RIGID \cite{rigid} does not perform well on certain facial datasets. To explore the root cause and motivated by robustness testing methods such as those used in ImageNet-C \cite{imagenetc}, we investigate whether different types of noise could enhance detection on these datasets. Notably, DINOv2 \cite{dinov2} has also highlighted the robustness of learned features through experiments on ImageNet-C \cite{imagenetc}.

\begin{figure}
    \centering
    \includegraphics[width=1\linewidth]{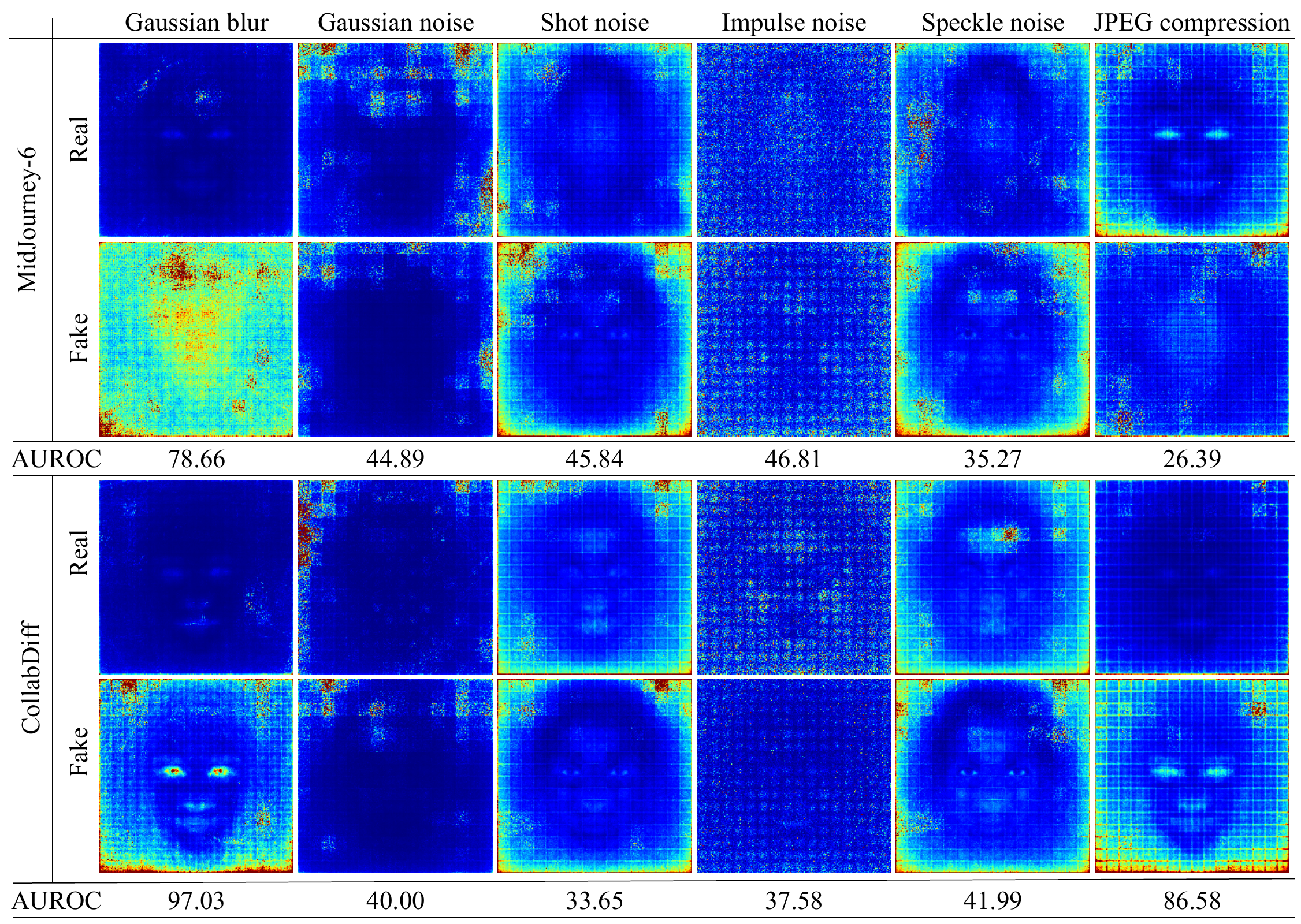}
    \caption{\textbf{Grad-CAM \cite{gradcam} of Feature Similarity with respect to Noise Across Different Perturbations.} Grad-CAM highlights the regions of interest (ROIs) in the output embedding of DINOv2, which are responsive to various types of added noise (at the pixel level). Note that we visualize the average Grad-CAM on facial images from MidJourney-6 \cite{midjourney} and CollabDiff \cite{collabdiff} datasets, respectively.}
    \label{fig:gradcam}
\end{figure}

We experimented with nearly all types of perturbations from ImageNet-C \cite{imagenetc}, and we found that Gaussian blur performs significantly better on deepfake images generated with MidJourney-6 \cite{midjourney} and CollabDiff \cite{collabdiff}. Specifically, a 3×3 kernel is generated using a Gaussian distribution, as shown below:
\begin{equation*}
     \text{kernel} (\sigma) = \frac{1}{\text{norm}}
\begin{bmatrix}
 f_{z}( \sqrt{2}) &  f_{z}(1) &  f_{z}( \sqrt{2}) \\
 f_{z}(1)         &  f_{z}(0) &  f_{z}(1) \\
 f_{z}( \sqrt{2}) &  f_{z}(1) &  f_{z}( \sqrt{2})
\end{bmatrix},
\end{equation*}
where $z \sim \mathcal{N}(0, \sigma^2)$ here and the kernel is normalized so that its $\ell_1$ norm equals 1. The kernel is then convolved with the image.

The test results are shown in Fig.~\ref{fig:gradcam}. Using Gaussian blur as the perturbation achieves an area under the receiver operating characteristic (AUROC) of 78.66, while other perturbations perform close to or worse than random guessing. The Grad-CAM \cite{gradcam} in the figure illustrates how each perturbation affects changes in the embeddings. For Gaussian noise, the noise is applied uniformly across the entire image. Interestingly, perturbations in the background contribute more to changes in embeddings than those in the foreground, making artifacts in facial regions harder to detect. In contrast, Gaussian blur effectively targets the facial region: the entire face generated by MidJourney-6 \cite{midjourney} is highly sensitive to this perturbation, and the eyes generated by CollabDiff \cite{collabdiff} are also notably affected. 

\subsection{Frequency Analysis}

\begin{figure}[t]
    \centering
    \includegraphics[width=1\linewidth]{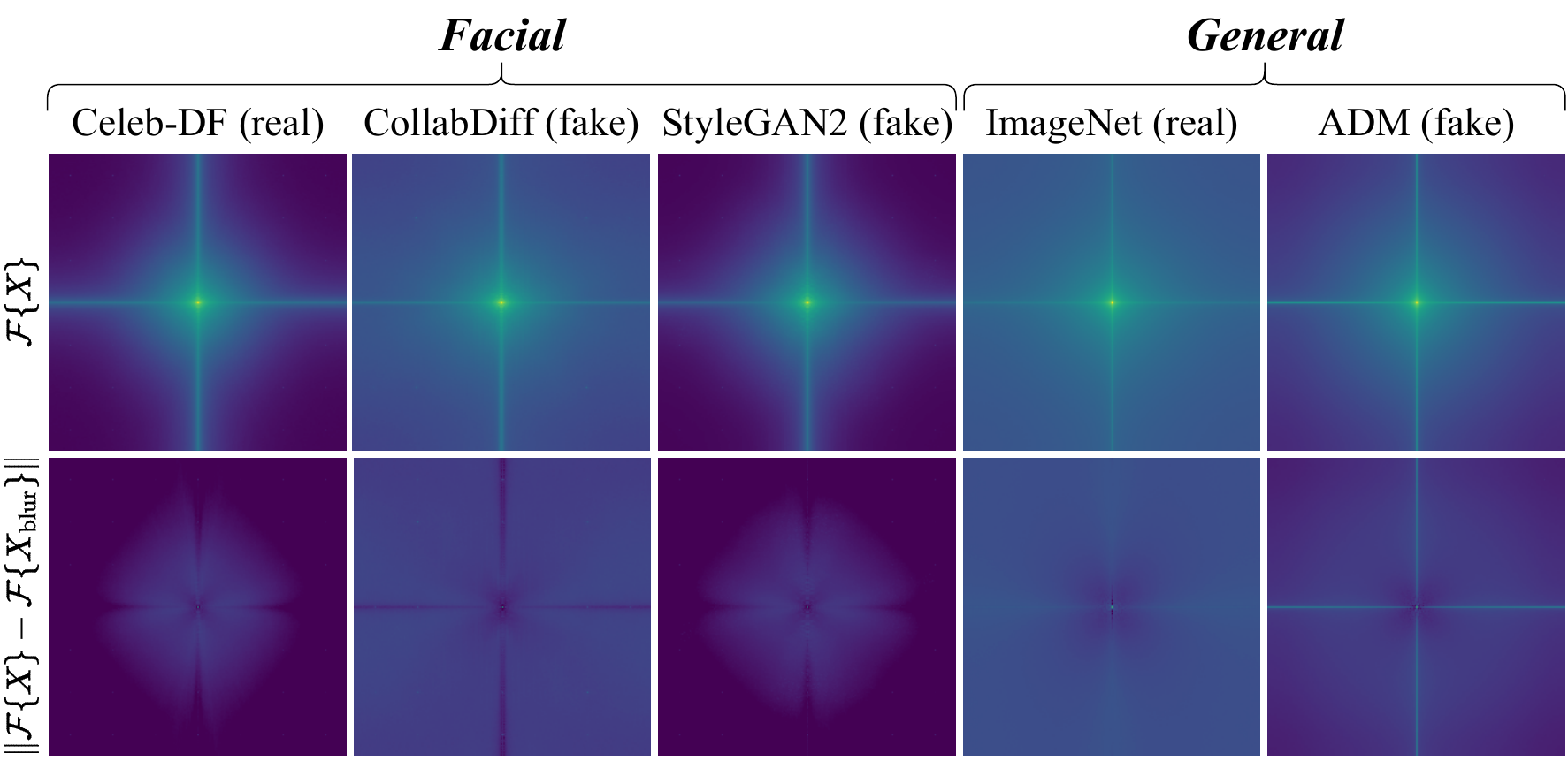}
    \caption{\textbf{Frequency spectrum visualization of real and fake images.} Images from CollabDiff \cite{collabdiff} and ImageNet \cite{imagenet} contain more high-frequency component, leading to greater perturbation. The notation $\mathcal{F}(\cdot)$ refers to the discrete Fourier transform and $X$ represents the image dataset. The frequency maps are center-shifted and  smoothed using $\log{(1 + x)}$ after being averaged over the entire dataset.}
    \label{fig:freq}
\end{figure}

In addition to the spatial regions that may trigger differences in embeddings, how does the information in the frequency domain contribute to these changes? While Gaussian blurring functions as a low-pass filter, its effectiveness may derive from the tendency of fake facial images to contain more high-frequency components than real images. While the noise introduced by blurring can be approximated with
\begin{equation*}
    \| \delta_{\text{blur}} \| = \| x - x_{\text{blur}} \| = \| x - \text{Filter}_{\text{low}}(x) \| \simeq \| x_{\text{high}} \|.
\end{equation*}
and, in general, larger noise leads to a greater distance between the original image embedding and the embedding of the perturbed image. From this, we can infer that a larger norm in the image’s high-frequency components causes a greater distance $\mathcal{D}(x, \delta)$, which in turn increases the likelihood of being detected as a fake image. A direct examination is provided in Table.~\ref{tab:l1}, where the distance function is replaced and the indicator function become 
\begin{equation*}
  y(x) = \textbf{1}\{ \ \| \delta_{\text{blur}}(x) \|_1 \geq \epsilon \ \}.
  \label{eq:l1}
\end{equation*}
The $\ell_1$ norm $\| \cdot \|_1 = \sum_{i, j} | \cdot_{ij} |$ denotes the sum of absolute values across all pixels. This method, when applied to facial images generated by CollabDiff \cite{collabdiff}, successfully classifies fake images based solely on the magnitude of the perturbation. However, detecting fake facial images cannot rely on this trivial approach alone. While the $\ell_1$ norm method performs similarly to random guessing when detecting fake images generated by StyleGAN2 \cite{sg2}, leveraging the robustness property of DINOv2 \cite{dinov2} enables successful detection of fake images. Interestingly, the trend of fake images containing more high-frequency components is reversed in the GenImage \cite{zhu2024genimage} dataset, but the robust nature of DINOv2 \cite{dinov2} still yields reasonable detection results. Frequency visualization is also provided in Fig.~\ref{fig:freq}. These observations suggest that the improved performance of the blurring method likely depends on both \textbf{frequency} artifacts and the \textbf{robust} feature extraction properties of DINOv2 \cite{dinov2}.

\subsection{Choice of Backbone Models}

\begin{table}[t]
    \centering
    \begin{tabular}{ccccc}  
         \toprule
         \multirow{2}{*}{Method}& \multicolumn{2}{c}{Facial} & \multicolumn{2}{c}{General} \\ \cmidrule(lr){2-3} \cmidrule(lr){4-5}
         & CollabDiff & StyleGAN2 & ADM & VQDM \\ \midrule
         $\| \delta_{\text{blur}} \|_1$ & \multicolumn{1}{c}{98.6} & 47.7 & 22.1 & 22.1 \\  
         $\mathcal{D}(x, \delta_{\text{blur}})$         & \multicolumn{1}{c}{97.0} & 93.3 & 64.0 & 81.2 \\ \bottomrule
    \end{tabular}
    \caption{\textbf{Experiments on potential frequency artifacts (AUROC score).} The first row indicate directly replacing $\mathcal{D}(f(x), f(x + \delta_{\text{blur}}))$ in the indicator function with $\|\delta_{\text{blur}}(x)\|_1$, meaning there is no need to pass images through the image encoder. This approach works for specific images from CollabDiff \cite{collabdiff}, the blurring method leverages the robustness of DINOv2 \cite{dinov2}, consistently improving performance.}
    \label{tab:l1}
\end{table}

The success of DINOv2 \cite{dinov2} inspired us to explore how it performs across different perturbations, encoder backbones, and datasets. To investigate whether these capabilities emerge suddenly in DINOv2, we included its predecessors, iBOT \cite{zhou2021ibot} and DINO \cite{dinov1}. We also considered different model sizes of DINOv2—base (B), large (L), and giant (g)—to examine whether there is a leap in performance when scaling up. Additionally, we tested other backbones that do not use self-supervised learning (SSL), including the vanilla ViT \cite{vit}, which is trained only with class labels, as well as contrastive learning models such as CLIP \cite{clip} and Nomic Embed Vision v1.5 \cite{nomic} trained with image-text pairs. As listed in Table~\ref{tab:backbone}, our findings and observation are:

\begin{itemize}
    \item \textbf{Finding 1: Self-supervised learning (SSL) enhances the robustness of learned representations and consistently outperforms non-SSL methods.} We observe that for effective perturbation-dataset pairs—namely, Gaussian noise applied to general images and Gaussian blur applied to facial images—SSL models significantly outperform non-SSL models. In these settings, SSL models demonstrate substantial performance gains, achieving improvements of up to 33.66 and 12.5, respectively. While DINOv2 \cite{dinov2} has demonstrated superior performance on the robustness testing dataset ImageNet-C \cite{imagenetc}, its performance in fake image detection is also better than that of other non-SSL methods. This further strengthens the claim that the effectiveness of perturbation-based methods stems from the model's robustness. Additionally, previous works, such as iBOT \cite{zhou2021ibot} and DINO \cite{dinov1}, show only slightly weaker performance compared to DINOv2, suggesting that this ability is not unique to DINOv2 and may extend to other models trained with SSL.

    \item \textbf{Finding 2: Performance gains through model scaling.} Results indicate that scaling up DINOv2 \cite{dinov2} improves performance on general images, particularly when transitioning from the base model to the large model (from 43.7 to 65.9) with Gaussian blur applied. The performance of the large and giant models is similar. These results align with the robustness scores reported in \cite{dinov2}, where the base, large, and giant models achieve scores of 42.7, 31.5, and 28.2 on ImageNet-C (lower is better), respectively, further demonstrating the relationship between detection performance and model robustness. For facial images, the performance improvement is minor. The frequency effects on facial images lead to effective performance, while Gaussian noise on facial images requires the help of DINOv2 for reliable detection.

    \item \textbf{Challenge: Detection performance is biased by perturbation selection.} Although Gaussian blur performs well on facial datasets, it introduces a performance bias that leads to significant degradation on general image datasets. We will discuss this challenge in more detail in the following section.
\end{itemize}

\begin{table}[h]
    \centering
    \begin{tabular}{ccccc}
        \toprule
         \multirow{2}{*}{Backbone} &  \multicolumn{2}{c}{Noise} & \multicolumn{2}{c}{Blur} \\ \cmidrule(lr){2-3} \cmidrule(lr){4-5}
         & Facial   & General   &  Facial  &   General \\ \midrule
         iBOT       & 46.8 & 82.4 & 93.0 & 37.8\\
         DINO       & 47.2 & 78.6 & 92.5 & 32.8\\
         DINOv2-B   & 51.7 & 85.0 & 96.7 & 43.7\\
         DINOv2-L   & 62.9 & 92.6 & 91.4 & 65.9\\
         DINOv2-g   & 66.7 & 90.5 & 91.8 & 72.6\\
         ViT        & 60.9 & 68.0 & 84.2 & 40.9\\
         CLIP       & 51.8 & 59.0 & 84.7 & 30.1\\
         Nomic 1.5  & 52.4 & 72.3 & 84.7 & 60.0\\ \bottomrule
    \end{tabular}
    \caption{\textbf{Experiments on Gaussian noise and Gaussian blur across different backbones (AUROC score).} The model's ability to provide robust features is positively correlated with performance. The facial score is averaged across the MidJourney-6, SD2.1, StyleGAN2, and CollabDiff datasets in DF40~\cite{df40}, while the general score is averaged across ADM and VQDM in the GenImage~\cite{zhu2024genimage} dataset.}
    \label{tab:backbone}
\end{table}
\section{Improving Training-Free Detection}
\label{sec:improve}

\begin{figure*}[t]
    \centering
    \includegraphics[width=0.865\linewidth]{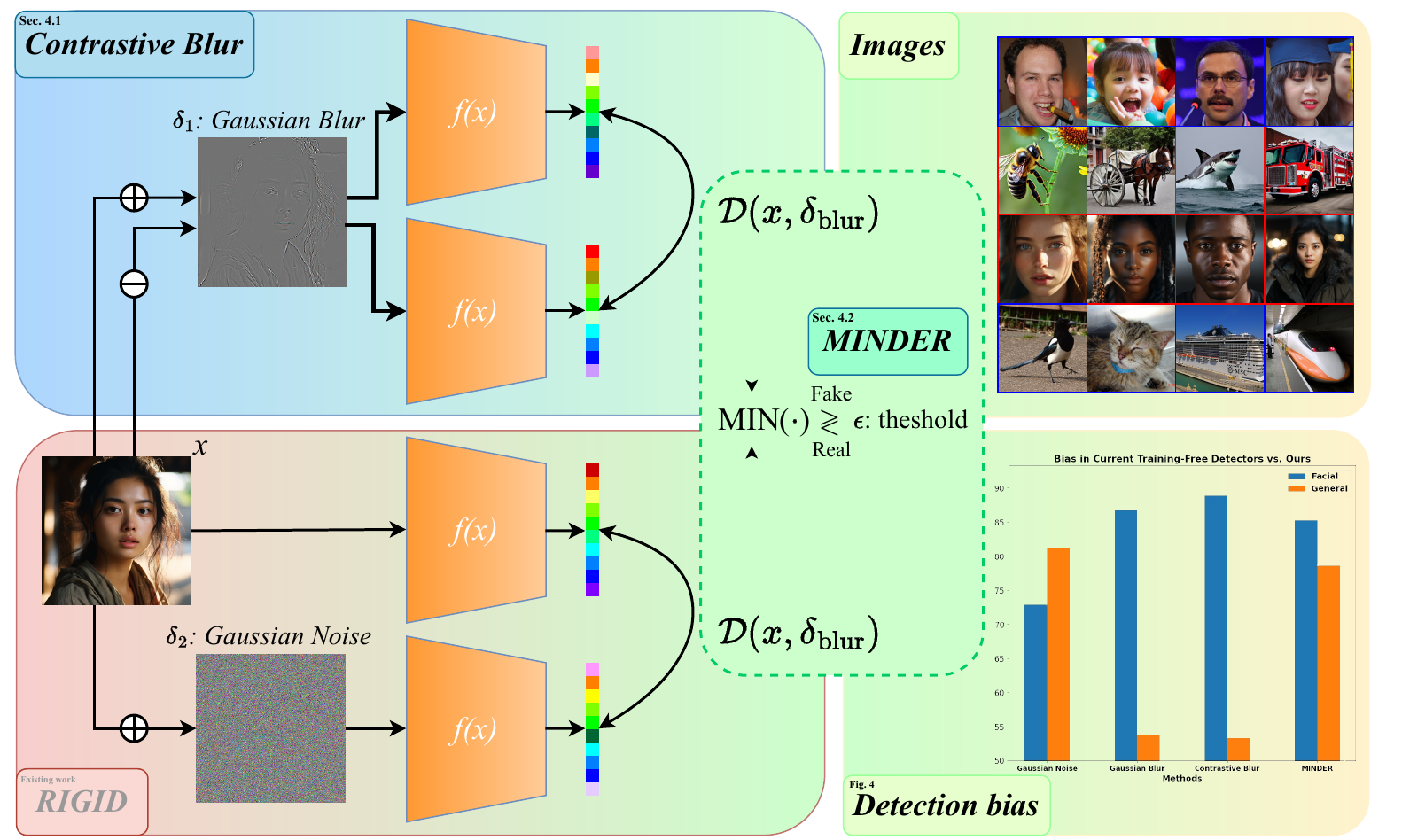}
    \caption{\textbf{Framework Overview.} Our framework extend the existing method RIGID \cite{rigid} by incorporating \textit{Contrastive Blur} (see Sec.\ref{sec:contrastive}) and \textbf{MINDER} (see Sec.~\ref{sec:minder}). This enhancement improves deepfake image detection performance and mitigates detection biases that arise from applying different types of noise. The results, shown in the bar plot, indicate a significant improvement in the average AUROC across two domains. In the figure, $f(x)$ denotes the vision encoder DINOv2. The fake images (the second and third rows) are also challenging to distinguish with the human eye.}
    \label{fig:fig1}
\end{figure*}

Based on the observations in the previous section, we found that frequency artifacts in deepfake images contribute to improved performance when the perturbation is switched from Gaussian noise to Gaussian blur. In this section, we introduce a new blurring-based method to further enhance the blurring operation. Additionally, the bias of specific noise types toward particular image types raises concerns. To address this, we propose \textbf{MINDER} (\textbf{MIN}imum distance \textbf{D}et\textbf{E}cto\textbf{R}) as a solution. The overall framework is illustrated in Fig.~\ref{fig:fig1}.

\subsection{Contrastive Blur}
\label{sec:contrastive}

While Gaussian blur demonstrates strong performance on deepfake datasets, we further explored methods to enhance its effectiveness. Unlike RIGID \cite{rigid}, which applies multiple samples of Gaussian noise to improve stability, the blurring process is deterministic and does not support multiple sampling. After unsuccessful trials with multi-level blurring, we found that calculating the distance between the embeddings of blurred and sharpened images yields better results. Specifically, 
\begin{equation*}
  y(x) = \textbf{1}\{ \ \mathcal{D} ( f(x + \delta_{\text{blur}}), \ f(x - \delta_{\text{blur}})) \geq \epsilon \ \} .
  \label{eq:contrastive}
\end{equation*}
We refer to this technique as \textit{Contrastive Blur} in the following sections. 
To motivate how contrastive blur helps us distinguish real and fake images, we see that while adding $\delta_{\text{blur}}$ reduces edge signals, subtracting $\delta_{blur}$ essentially sharpens the edges in an image. Now if we express the distance induced by contrastive blur ($D ( f(x + \delta_{\text{blur}}), f(x - \delta_{\text{blur}}))$) by $D(x, \delta_{\text{blur}})$ and $D(x, -\delta_{\text{blur}})$ for both real and fake images, we can evaluate how real and fake images react to blurring/sharpening. That is, if we express $D ( f(x + \delta_{\text{blur}}), f(x - \delta_{\text{blur}}))$ by $\alpha D(x, \delta_{\text{blur}}) + \beta D(x, -\delta_{\text{blur}}) + \gamma$ for $x$ being real or fake images, experiments on StyleGANXL \cite{sgxl} receive $\alpha=0.79,\beta=1.29, \gamma=3.0 \times 10^{-4}$ for real images, and $\alpha=1.66, \beta=1.45, \gamma=1.5 \times 10^{-4}$ for fake images, suggesting that fake facial images experience a stronger linearity in distance addition, thus further increase the distance. Therefore, we can leverage this observation to enhance the fake facial image detection. Further visualization is provided in the supplementary material.

\subsection{Minimum Distance Detector}
\label{sec:minder}

\begin{figure}[h]
    \centering
    \includegraphics[width=1\linewidth]{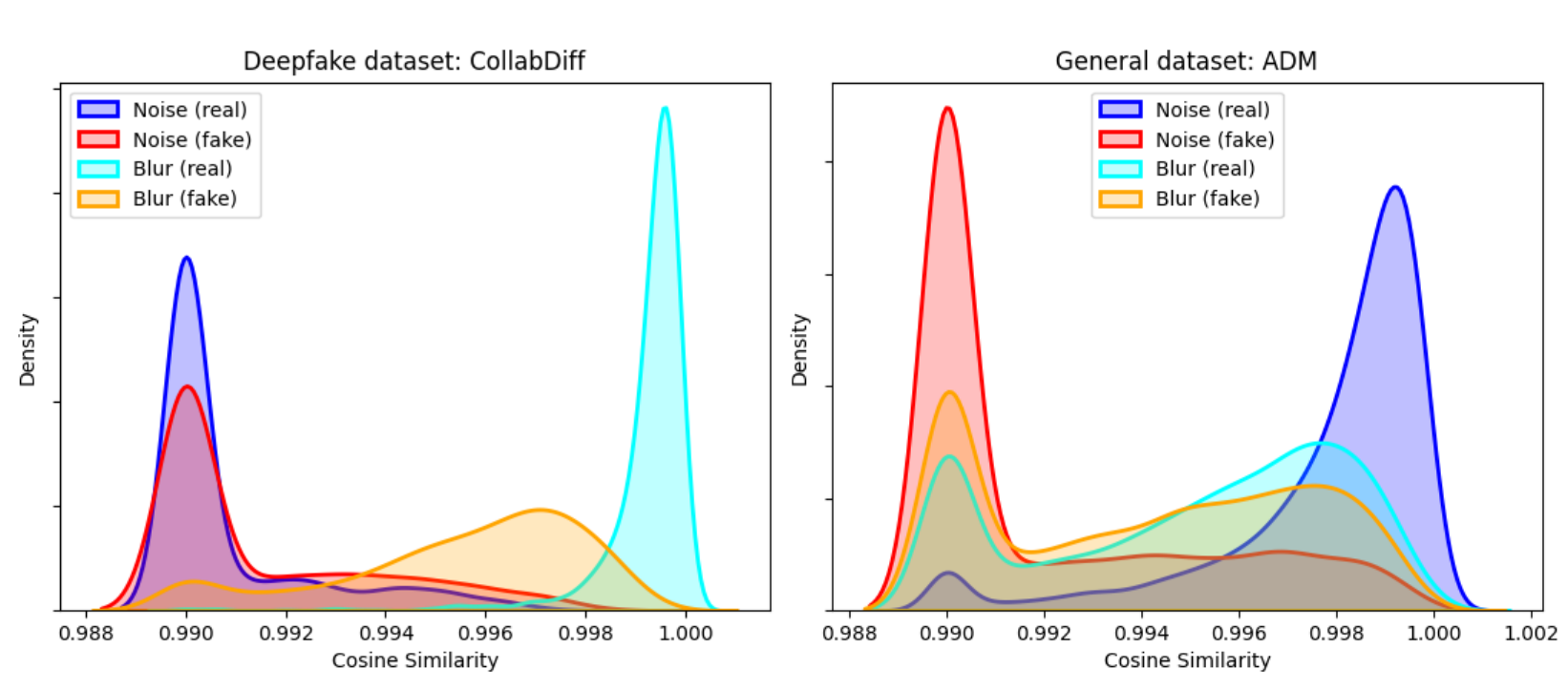}
    \caption{\textbf{Distribution Visualization of Cosine Similarity after Perturbations.} Blurring produces a gap in the cosine similarity distribution for facial images generated by CollabDiff \cite{collabdiff}, while for general images in ADM \cite{adm}, the distribution becomes indistinguishable. This trend reverses when applying Gaussian noise.}
    \label{fig:dist}
\end{figure}

While Gaussian blur shows significant improvement on the deepfake dataset, it unfortunately does not work as well on general images. As shown in Fig.~\ref{fig:dist}, while real and fake images in the deepfake dataset CollabDiff \cite{collabdiff} exhibit different sensitivities to Gaussian blur, real and fake general images generated by ADM \cite{adm} show similar sensitivity to Gaussian blur. To address this issue, we explore a new method that leverages different types of noise to achieve robust performance across both facial and general domains.

Based on the observation in Fig.~\ref{fig:dist}, we propose \textbf{MINDER} using the \textbf{MIN}imum distance to balance performance between facial and general images. The result in Fig.~\ref{fig:dist} demonstrate that fake images are relatively more sensitive to both Gaussian noise ($\delta_1$) and Gaussian blur ($\delta_2$) when considering the worst-case scenario. Formally, let us consider two metric spaces $(X, d_1)$ and $(X, d_2)$ over the set of all images $X$, and a constant threshold $\epsilon > 0$. We are interested in two different scenarios, where
\begin{equation}
\begin{aligned}
    A &= \{ (x, y) \in X \times X : d_1(x, y) > \epsilon \}, \\
    B &= \{ (x, y) \in X \times X : d_2(x, y) > \epsilon \}.
\end{aligned}
\label{eq:minder}
\end{equation}
In particular, we are interested in $A \cup B$, which is
\begin{equation*}
    \{ (x, y) \in X \times X : \min(d_1(x, y), d_2(x, y)) > \epsilon \}.
\end{equation*} This operation can be interpreted by the fact that real images are at least robust to one of the two types of noise: Gaussian noise or Gaussian blur. By selecting the minimum distance, we produce a low threshold that allows only real images to pass through. The distance function in the indicator function then becomes
\begin{equation}
    \mathcal{D}_{\text{min}}(\cdot) = \min(\mathcal{D}(x, \delta_1), \mathcal{D}(x, \delta_2)).
    \label{eq:dmin}
\end{equation}
This method successfully addresses the detection bias evident in the bar plot in Fig.~\ref{fig:fig1}.

Other methods, such as \textbf{MIX}ing noise in the image space via arithmetic addition, are also considered in the experiments: 
\begin{equation*}
    \delta_{\text{mix}} = \delta_1 + \delta_2.
\end{equation*}
As an ablation study, we also consider the \textbf{MAX}imum distance for evaluation: 
\begin{equation*}
    \mathcal{D}_{\text{max}}(\cdot) = \max(\mathcal{D}(x, \delta_1), \mathcal{D}(x, \delta_2)).
\end{equation*}

\begin{table*}[ht]
    \centering
    \begin{tabular}{cccccccc}
        \toprule
    \multirow{2}{*}{Dataset} & \multicolumn{3}{c}{Training based}    & \multicolumn{4}{c}{Training free} \\ \cmidrule(lr){2-4} \cmidrule(lr){5-8}
                & NPR   & DF40  & AIDE  & AeroBlade & RIGID & Ours (blur) & Ours (contrastive)\\ \midrule
HeyGen          & 73.3 & 86.8 & 31.6 & 21.1 & 28.7 & \textbf{91.8} & \underline{90.1} \\ 
DDIM            & \textbf{98.6} & 75.4 & \underline{93.8} & 2.76 & 78.1 & 89.3 & 92.4 \\ 
DiT             & 21.1 & \underline{80.5} & \textbf{88.0} & 98.3 & 58.8 & 70.5 & 76.6 \\ 
MidJourney-6    & 29.6 & \underline{87.2} & 34.0 & \textbf{94.8} & 44.6 & 78.7 & 87.0 \\
PixArt-$\alpha$ & \textbf{100}  & 89.6 & 64.8 & \underline{96.1} & 90.7 & 87.8 & 82.3 \\
SD2.1           & \textbf{99.7} & 75.6 & 93.8 & 74.9 & 79.9 & \underline{96.5} & 96.3 \\ 
SiT             & \textbf{99.2} & 81.2 & 89.8 & \underline{97.4} & 61.5 & 72.8 & 78.5 \\ 
StyleGAN2       & \textbf{100}  & 88.1 & 76.7 & 24.0 & 88.8 & \underline{93.3} & 91.2 \\ 
StyleGAN3       & \textbf{100}  & 92.2 & 80.4 & 77.3 & \underline{98.7} & 93.3 & 93.0 \\ 
StyleGAN-XL     & \textbf{100}  & \underline{95.7} & 75.7 & 34.7 & 81.6 & 73.2 & 81.3 \\
VQGAN           & \textbf{99.9} & 95.8 & 85.3 & 43.1 & 94.3 & 99.0 & \underline{99.4} \\
Whichisreal     & 54.7 & \underline{85.5} & 39.9 & 50.5 & \textbf{88.3} & 79.3 & 83.7 \\
CollabDiff      & \textbf{100}  & 54.1 & 33.1 & 2.16 & 41.6 & 97.0 & \underline{98.3} \\ 
e4e             & 91.5 & 72.8 & 64.2 & 29.7 & 96.2 & \underline{97.7} & \textbf{98.9} \\
StarGAN2        & 49.3 & \textbf{90.9} & 54.8 & 36.9 & \underline{76.0} & 68.9 & 73.3 \\
StyleCLIP       & 90.0 & 92.9 & 96.9 & 55.6 & 68.0 & \textbf{98.5} & \underline{98.1} \\ \midrule
Avg.            & \textbf{90.5} & 84.0 & 67.5 & 52.5 & 73.5 & 86.7 & \underline{88.8} \\ \bottomrule
    
    \end{tabular}
    \caption{\textbf{Experiment on the Deepfake Dataset DF40 \cite{df40} (AUROC score).} Our method outperforms the state-of-the-art training-free approach and achieves performance comparable to training-based methods. The best result are in \textbf{bold text} and the second-best results are \underline{underlined}. For a fair comparison to RIGID \cite{rigid}, we set $n=3$, sample three Gaussian noise instances, and calculate the mean distance.}    \label{tab:main}
\end{table*}
\section{Experiments}
\label{sec:exp}

\subsection{Setup}

\paragraph{Datasets.} We evaluate the performance on the subsets of deepfake dataset DF40 \cite{df40}, which contains real images from Celeb-DF (CDF) \cite{cdf}, FFHQ \cite{ffhq} and CelebA \cite{celeba}, as well as images generated by various generation and editing methods: HeyGen \cite{HeyGen}, DDIM \cite{ddim}, DiT \cite{dit}, MidJourney-6 \cite{midjourney}, PixArt-$\alpha$ \cite{pixart}, Stable Diffusion v2.1 \cite{sd}, SiT \cite{sit}, StyleGAN2 \cite{sg2}, StyleGAN3 \cite{sg3}, StyleGAN-XL \cite{sgxl}, VQGAN \cite{vqgan}, Whichisreal \cite{whichisreal}, CollabDiff \cite{collabdiff}, e4e \cite{e4e}, StarGAN2 \cite{starganv2}, and StyleCLIP \cite{styleclip}. We also test on the general images dataset GenImage \cite{zhu2024genimage}, which contains real images from ImageNet \cite{imagenet} and numerous generation methods such as ADM \cite{adm}, BigGAN \cite{biggan}, Glide \cite{nichol2021glide}, MidJourney \cite{midjourney}, Stable Diffusion v1.4 and v1.5 \cite{sd}, VQDM \cite{vqdm}, and Wukong \cite{wukong}. The experiments are conducted on the test set of DF40 \cite{df40} and the validation set of GenImage \cite{zhu2024genimage}, with all images resized to 224x224. 


\paragraph{Metrics.} We evaluate performance using the area under the receiver operating characteristic curve (AUROC), following the metric used in previous works


\paragraph{Baselines.} We select several state-of-the-art methods for deepfake and AI-generated image detection, including both training-based and training-free approaches. For training-based methods, NPR \cite{npr} is trained on fake images generated by ProGAN \cite{progan} and real images from LSUN \cite{yu2015lsun}; AIDE \cite{aide} is trained on fake images generated by Stable Diffusion v1.4 \cite{sd} and real images from LSUN \cite{yu2015lsun}. The CLIP model in DF40 \cite{df40} is trained on real images from FaceForensics++ \cite{rossler2019faceforensics++} and fake images from the DF40 \cite{df40} training set. For training-free methods, AeroBlade \cite{ricker2024aeroblade} uses a reconstruction-based approach, assuming that images generated by a latent diffusion model will exhibit smaller reconstruction errors after encoding and decoding operations. RIGID \cite{rigid} exploits the sensitivity of fake images to Gaussian noise perturbation. Unfortunately, ZED \cite{zed} cannot release its trained weights due to funding constraints, although it shows strong performance in detecting AI-generated images.

\subsection{Evaluation} 

\paragraph{Main Results.} Table~\ref{tab:main} demonstrates that our method performs favorably against training-free and learning-based methods on subsets of DF40 \cite{df40}, with a \textbf{15.3 point improvement} in AUROC. Notable improvements are observed in the MidJourney-6 \cite{midjourney} and DiT \cite{dit} datasets.

\paragraph{Results of Contrastive Blur.} The results in Table.~\ref{tab:contrastive} show that while blurring alone is sensitive to parameter changes, with performance ranging from 83.8 to 87.3, \textit{Contrastive Blur} consistently improves performance by 0.32 to 4.07.

\begin{table}[h]
    \centering
    \begin{tabular}{cccc}
        \toprule
                & $\sigma=0.45$ & $\sigma=0.55$ & $\sigma=0.65$ \\  \midrule

Blur                &  87.29 & 86.73 & 83.77\\
\textit{Contrastive Blur}    & \textbf{87.61} & \textbf{88.77} & \textbf{87.84}\\ \bottomrule

    \end{tabular}
    \caption{\textbf{Experiments on Contrastive Blur with the deepfake dataset DF40 \cite{df40} (AUROC score).} \textit{Contrastive Blur} consistently improves performance, pushing the score above 87.5.}
    \label{tab:contrastive}
\end{table}

\vspace{-0.5em}

\paragraph{Results of MINDER.} As shown in Table~\ref{tab:ensemble}, the \textbf{MIN}imum distance strategy effectively maintains performance on DF40 \cite{df40} and GenImage \cite{zhu2024genimage}, with minor AUROC drops of 1.4 and 2.6, respectively, compared to the best method. This results in a 4.9 point improvement in the average score. Other operations \textbf{MIX} and \textbf{MAX} seem to favor facial images.

\begin{table}[h]
    \centering
    \begin{tabular}{cccccc}
        \toprule
                & Noise & Blur & MIX & MAX & MIN \\  \midrule

DF40        & 72.9 & \textbf{86.7} & 81.0 & 78.7 & \underline{85.3} \\
GenImage    & \textbf{81.2} & 53.9 & 65.2 & 66.6 & \underline{78.6} \\ \midrule
Avg.        & \underline{77.1} & 70.3 & 73.1 & 72.7 & \textbf{82.0} \\ \bottomrule

    \end{tabular}
    \caption{\textbf{Experiments on MINDER with deepfake dataset DF40 \cite{df40} and general dataset GenImage \cite{zhu2024genimage} (AUROC score).} The \textbf{MIN} operation successfully leverages both Gaussian noise and Gaussian blur, while the \textbf{MIX} and \textbf{MAX} operations do not yield effective results.}
    \label{tab:ensemble}
\end{table}

\vspace{-0.5em}

\paragraph{Ablation Studies.} The \textit{Contrastive Blur} technique improves DF40 \cite{df40} performance by 2.1, while the ensemble strategy boosts the average score by 9.1. Combining both raises the improvement to 11.7, demonstrating the effectiveness of each component's incremental contribution.

\begin{table}[ht]
    \centering
    \begin{tabular}{lccc}
        \toprule
        Method & \hspace{0.8em} DF40 \hspace{0.8em} & GenImage & Avg. \\  \midrule
        Blur          & 86.7 & 53.9 & 70.3 \\
        + Contrastive & 88.8 & 53.3 & 71.1 (+0.8) \\ 
        + MINDER      & 85.1 & 73.7 & 79.4 (+9.1)\\ 
        + Both        & 85.3 & 78.6 & 82.0 (+11.7)\\ \bottomrule
    \end{tabular}
    \caption{\textbf{Ablation study on each component (AUROC score).} Consistent improvements demonstrate the effectiveness of each component.}
    \label{tab:ablation}
\end{table}
\section{Ongoing Challenges and Limitations}
\label{sec:limitation}

\paragraph{Ongoing Challenges.} Although our method has demonstrated significant improvements on certain deepfake datasets, the perturbation-based approach, which measures global embedding sensitivity, operates as a statistic over the entire image. This approach is particularly effective for forgery methods that substantially alter the entire image, such as StyleCLIP \cite{styleclip}, which optimizes the entire image to align with text embeddings. However, for face swapping (FS) and face reenactment (FR) datasets, this method struggles to detect even obvious boundary artifacts. This limitation arises because it relies on global information, which can obscure important local details. 

\begin{figure}[h]
    \centering
    \includegraphics[width=1\linewidth]{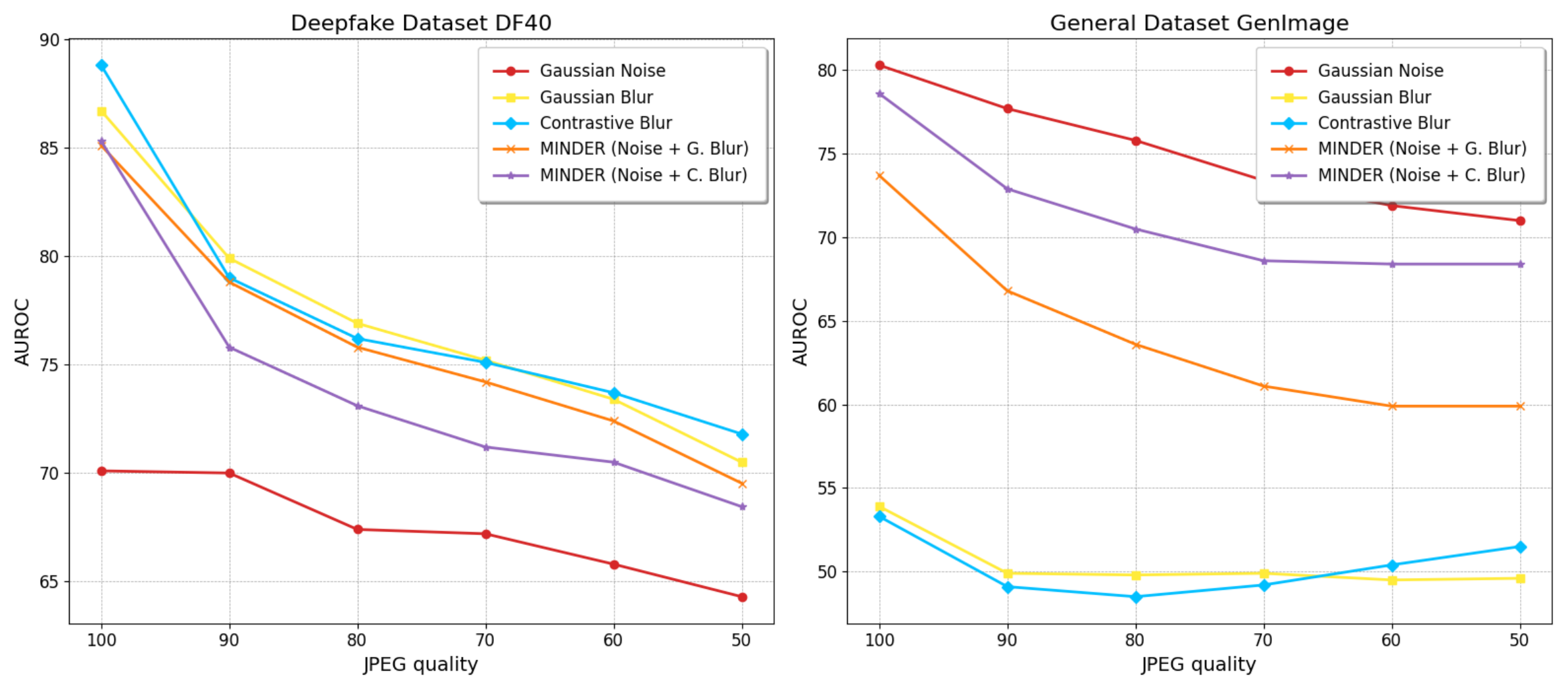}
    \caption{\textbf{Performance degradation under JPEG compression.} As expected, Gaussian blur is more sensitive to frequency corruption, leading to a larger decrease in performance. However, \textbf{MINDER} remains robust in effectively leveraging both types of noise, even though the individual methods it combines are affected by JPEG compression.}
    \label{fig:jpeg}
\end{figure}

For FS and FR, the average AUC scores are 52.88 and 54.83, respectively—values close to random guessing. While attempts to capture local distributions have been made, addressing this within the perturbation framework remains a challenging task.
\paragraph{Limitations.} While the blurring method depends on frequency distribution, its effectiveness under JPEG compression warrants further investigation. JPEG compression quantizes discrete Fourier transform coefficients, typically discarding high-frequency components. As shown in Fig.~\ref{fig:jpeg}, the blurring method’s performance declines significantly, as anticipated. In contrast, \textbf{MINDER} maintains its effectiveness by leveraging both noise types. Expanding our framework to handle additional noise types is a key research direction. Additionally, performance variations across different generation methods present an ongoing challenge for the detection community.

\section{Conclusion}
\label{sec:conclusion}

This paper presents a comprehensive analysis of the perturbation-based training-free AI-generated image detection methods. We provide new insights in examining frequency characteristics and detection biases, proposing an enhanced framework with Contrastive Blur and \textbf{MINDER} to improve detection performance for facial images and generalize across facial and general datasets. Extensive experiments confirm the effectiveness of our methods in boosting detection accuracy. Our training-free method achieves the best training-free detection performance and is even on par with training-based approaches. We believe that exploring training-free detection and identifying key statistical properties provides valuable insights for advancing research in image generation and detection.

{
    \small
    \bibliographystyle{unsrtnat}
    \bibliography{main}
}

\clearpage
\setcounter{page}{1}
\maketitlesupplementary

\setcounter{section}{0}
\renewcommand{\thesection}{\Alph{section}}

\section{Visualization of Contrastive Blur}

\begin{figure*}[ht]
    \centering
    \includegraphics[width=1\linewidth]{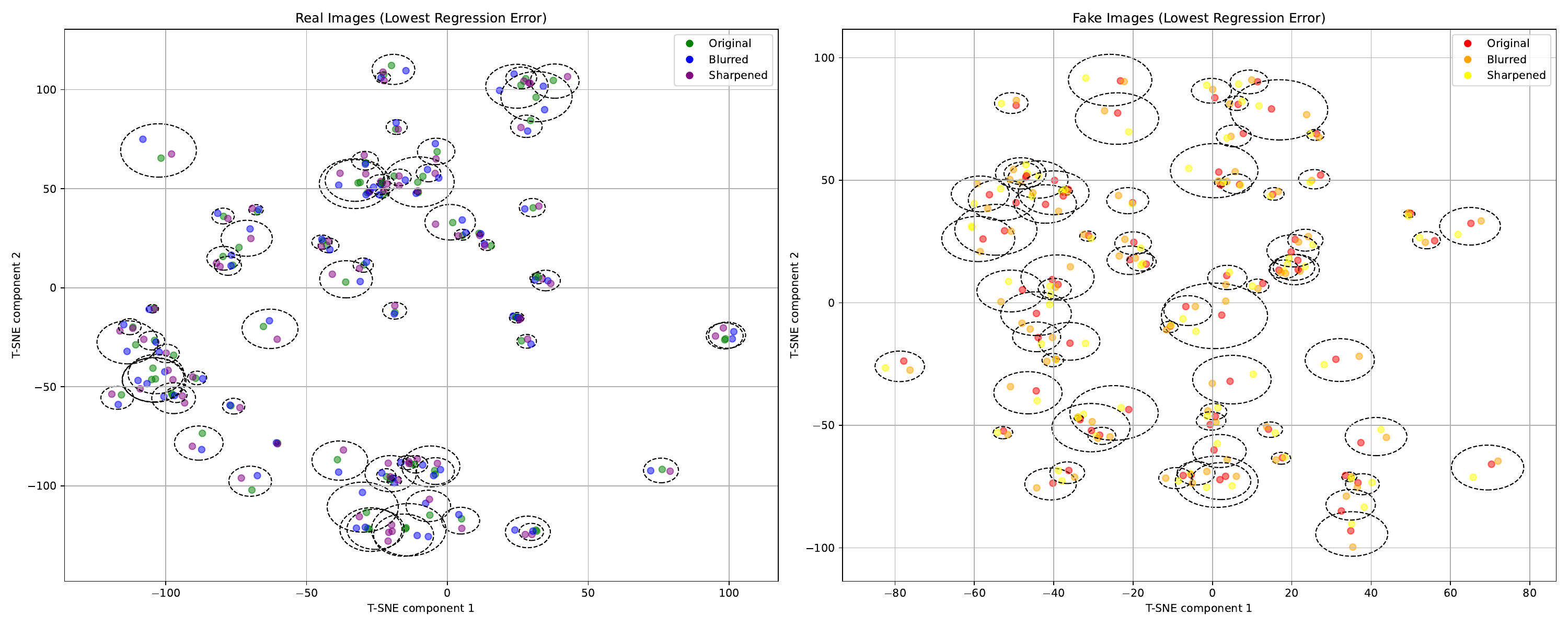}
    \caption{\textbf{T-SNE visualization of Contrastive Blur.}}
    \label{fig:tsne}
\end{figure*}

We provide a visualization to illustrate the technique described in Sec.~\ref{sec:contrastive}. In Fig.~\ref{fig:tsne}, we select 100 data points with the lowest regression errors and scale the differences between original embeddings and their blurred or sharpened counterparts for better visualization. Embeddings corresponding to the same image are grouped in a circle. The figure shows that the shifts caused by blurring and sharpening are often in opposite directions, creating a relationship like $\text{distance}_{\text{contrastive}} = \text{distance}_{\text{blurred}} + \text{distance}_{\text{sharpen}}$. Interestingly, for real images, the shifts caused by blurring and sharpening tend to align in the same or similar directions, often with an angle less than 90 degrees.

\section{Implementation Details}

All perturbations are applied in the image space, with pixel values normalized to the range $[0, 1]$. Most implementations are based on the code provided by \cite{df40} and ImageNet-C \cite{imagenetc}. The Gaussian blur with a 3x3 kernel is implemented as part of the defocus blur corruption applied to CIFAR images. After perturbation, the images are clipped to the range $[0, 1]$. All experiments are conducted on a single NVIDIA Titan RTX GPU. In the MINDER experiments, we use $\sigma_{\text{blur}}=0.009$, $\sigma_{\text{blur}}=0.55$ and $\mathcal{D} = \mathcal{D}_{\text{min}}$ in Eq.~\ref{eq:dmin}.

\section{DF40 Datasets.}

Table~\ref{tab:dataset} summarizes the model types of forgery methods (GAN or diffusion) along with their venues.

We report the performance of forgery detection methods across three categories: entire face synthesis (EFS), face editing (FE), and the face reenactment dataset HeyGen \cite{HeyGen}. The RDDM dataset \cite{rddm} was excluded because the fake images it generated contained excessive noise, resulting in an anomalously low AUC of 0.2 for Gaussian noise perturbation. In contrast, our blurring method achieved a high AUC of 99.9. Furthermore, the StarGAN \cite{df40} subset in the FE category is missing from the dataset provided by the DF40 authors \cite{df40}. Partial results for face swapping (FS) and face reenactment (FE) are presented in Table~\ref{tab:fsfe}.

\section{Details of Multi-Noise Strategy}

\paragraph{MIX.} If two perturbations are independent and do not adversely affect the robustness property, they can be combined directly in the image space:
\begin{equation*}
    \delta_{\text{mix}} = \mathrm{Agg}(\delta_1, \delta_2, \dots, \delta_n),
\end{equation*}
where $\mathrm{Agg}(\cdot)$ represents an aggregation function, such as arithmetic addition. Given the conditions:
\begin{equation*}
    \mathcal{D}(x_{\text{facial fake}}, \delta_1) \geq \epsilon_1, \quad 
    \mathcal{D}(x_{\text{general fake}}, \delta_2) \geq \epsilon_2
\end{equation*}
and the approximations:
\begin{equation*}
\begin{aligned}
    \mathcal{D}(x, \delta_1 + \delta_2) &\simeq \mathcal{D}(x, \delta_1) + \mathcal{D}(x, \delta_2), \\
    \mathcal{D}(x_{\text{facial fake}}, \delta_2) &\simeq \mathcal{D}(x_{\text{facial real}}, \delta_2) = c_1, \\
    \mathcal{D}(x_{\text{general fake}}, \delta_1) &\simeq \mathcal{D}(x_{\text{general real}}, \delta_1) = c_2, \\
\end{aligned}
\end{equation*}
fake images can still be detected using a threshold:
\begin{equation*}
    \mathcal{D}(x_{\text{fake}}, \delta_1 + \delta_2) \geq \min(\epsilon_1, \epsilon_2) + (c_1 + c_2).
\end{equation*}
However, the increased threshold slightly raises the false negative rate.

\paragraph{MAX.} Assuming that fake images are particularly sensitive to a \textbf{specific} type of noise within a set, and following the formulation in \textbf{MIN}~\ref{eq:minder}, the set $A \cap B$ becomes
\begin{equation*}
    \{ (x, y) \in X \times X : \max(d_1(x, y), d_2(x, y)) > \epsilon \}.
\end{equation*}
Accordingly, the distance function is:
\begin{equation*}
    \mathcal{D}_{\text{max}}(\cdot) = \max(\mathcal{D}(x, \delta_1), \mathcal{D}(x, \delta_2)).
\end{equation*}

\section{Guassian Kernels Values}
\begin{equation*}
     \text{kernel} (\sigma) = \frac{1}{\text{norm}}
\begin{bmatrix}
 f_{z}( \sqrt{2}) &  f_{z}(1) &  f_{z}( \sqrt{2}) \\
 f_{z}(1)         &  f_{z}(0) &  f_{z}(1) \\
 f_{z}( \sqrt{2}) &  f_{z}(1) &  f_{z}( \sqrt{2})
\end{bmatrix},
\end{equation*}
\begin{equation*}
\begin{aligned}
\text{kernel} (0.45) &= 
\begin{bmatrix}
0.005 & 0.062 & 0.005 \\
0.062 & 0.731 & 0.062 \\
0.005 & 0.062 & 0.005
\end{bmatrix}; \\
\text{kernel} (0.50) &= 
\begin{bmatrix}
0.011 & 0.084 & 0.011 \\
0.084 & 0.619 & 0.084 \\
0.011 & 0.084 & 0.011
\end{bmatrix}; \\
\text{kernel} (0.55) &= 
\begin{bmatrix}
0.019 & 0.100 & 0.019 \\
0.100 & 0.523 & 0.100 \\
0.019 & 0.100 & 0.019
\end{bmatrix}; \\
\text{kernel} (0.60) &= 
\begin{bmatrix}
0.028 & 0.111 & 0.028 \\
0.111 & 0.445 & 0.111 \\
0.028 & 0.111 & 0.028
\end{bmatrix}; \\
\text{kernel} (0.65) &= 
\begin{bmatrix}
0.036 & 0.118 & 0.036 \\
0.118 & 0.385 & 0.118 \\
0.036 & 0.118 & 0.036
\end{bmatrix}.
\end{aligned}
\end{equation*}

\section{Extensive Experiments with Different Parameters and Detailed Results}

Detailed results of selecting different standard deviations for the Gaussian blur kernel are listed in Table~\ref{tab:blur_full}. Detailed results of selecting different standard deviations for contrastive blur are shown in Table~\ref{tab:contrastive_full}. The necessary balance between the strengths of the two noise types is displayed in Table~\ref{tab:ensemble_suppl}. Detailed results of the ensemble strategy are listed in Table~\ref{tab:ensemble_full}.

\begin{table*}
    \centering
    \begin{tabular}{cccc}
        \toprule
                 & Dataset & Model type & Venue\\ \midrule
        \multirow{15}{*}{\rotatebox{90}{DF40}}
        &HeyGen          & Not opened       & None\\ 
        &DDIM            & Diffusion        & ICLR 2021\\ 
        &DiT             & Latent diffusion & ICCV 2023\\ 
        &MidJourney-6    & Not opened       & None\\
        &PixArt-$\alpha$ & Latent diffusion & ICLR 2024\\
        &SD2.1           & Latent diffusion & CVPR 2022\\ 
        &SiT             & Latent diffusion & Arxiv 2024\\ 
        &StyleGAN2       & GAN              & Arxiv 2019\\ 
        &StyleGAN3       & GAN              & NeurIPS 2021\\ 
        &StyleGAN-XL     & GAN              & SIGGRAPH 2022\\
        &VQGAN           & GAN              & CVPR 2021\\
        &Whichisreal     & GAN              & None\\
        &CollabDiff      & Latent diffusion & CVPR 2023\\ 
        &e4e             & GAN              & SIGGRAPH 2021\\
        &StarGAN2        & GAN              & CVPR 2020\\
        &StyleCLIP       & GAN              & ICCV 2021\\ \bottomrule \toprule
        \multirow{8}{*}{\rotatebox{90}{GenImage}}
        &ADM             & Diffusion        & NeurIPS 2021\\
        &BigGAN          & GAN              & ICLR 2019\\
        &Glide           & Diffusion        & Arxiv 2021 \\
        &MidJourney      & Not opened       & None\\
        &SD1.4           & Latent Diffusion & CVPR 2022\\
        &SD1.5           & Latent Diffusion & CVPR 2022\\
        &VQDM            & Latent Diffusion & CVPR 2022\\
        &Wukong          & Latent Diffusion & None\\ \bottomrule
    \end{tabular}
    \caption{\textbf{Dataset Details: Model Type and Venue.} The content is partially derived from DF40 \cite{df40}.
}
    \label{tab:dataset}
\end{table*}

\begin{table*}
    \centering
    \begin{tabular}{cccccc}
        \toprule
        Dataset         & NPR & DF40 & Aeroblade & RIGID & Blur \\ \midrule
        FSGAN           & 48.3 & 96.1 & 51.9 & 48.6 & 47.1 \\ 
        FaceSwap        & 47.4 & 98.1 & 45.6 & 53.2 & 53.2 \\ 
        SimSwap         & 65.8 & 91.4 & 51.0 & 45.3 & 49.9 \\ 
        InSwapper       & 64.0 & 82.3 & 55.4 & 50.1 & 50.2 \\
        BlendFace       & 63.4 & 94.9 & 55.6 & 44.4 & 49.3 \\
        UniFace         & 73.4 & 94.6 & 61.8 & 45.2 & 55.7 \\ 
        MobileSwap      & 45.5 & 93.5 & 42.4 & 51.0 & 52.8 \\ 
        e4s             & 35.7 & 97.7 & 12.2 & 58.1 & 67.1 \\ 
        FaceDancer      & 14.2 & 97.4 & 27.2 & 49.3 & 50.7 \\ 
        DeepFaceLab     & 35.5 & 94.7 & 61.4 & 53.5 & 53.0 \\ \midrule
        Avg.            & 49.3 & 94.1 & 46.5 & 49.9 & 52.9 \\ \bottomrule \toprule
        FOMM            & 64.5 & 94.2 & 61.0 & 77.8 & 56.6 \\
        FSvid2vid       & 39.0 & 91.6 & 60.6 & 73.8 & 47.1 \\
        Wav2Lip         & 56.9 & 83.3 & 49.4 & 41.7 & 44.3 \\ 
        MRAA            & 73.7 & 94.5 & 82.5 & 76.4 & 49.5 \\
        OneShot         & 63.5 & 92.6 & 58.4 & 77.8 & 53.3 \\
        PIRender        & 66.6 & 94.0 & 58.0 & 71.0 & 50.8 \\ 
        TPSMM           & 40.9 & 84.1 & 53.9 & 52.6 & 55.6 \\
        LIA             & 78.4 & 86.2 & 71.5 & 70.3 & 61.9 \\
        DaGAN           & 38.8 & 90.0 & 65.0 & 72.0 & 52.3 \\
        SadTaker        & 37.9 & 79.5 & 54.3 & 51.3 & 49.2 \\
        MCNet           & 39.3 & 90.1 & 69.1 & 73.8 & 54.1 \\
        HyperReenact    & 57.1 & 95.3 & 59.5 & 72.9 & 46.3 \\
        HeyGen          & 73.3 & 86.8 & 21.1 & 29.6 & 91.8 \\ \midrule        
        Avg.            & 56.1 & 89.4 & 58.8 & 64.7 & 54.7\\ \bottomrule
    \end{tabular}
    \caption{\textbf{Experiments on face swapping (FS) and face reenactment (FR).} While the DF40 \cite{df40} CLIP detector continues to perform effectively, the performance of other methods degrades significantly.}
    \label{tab:fsfe}
\end{table*}

\begin{table*}
    \centering
    \begin{tabular}{cccccc}
        \toprule
        Dataset         & $\sigma=0.45$ & $\sigma=0.50$ & $\sigma=0.55$ & $\sigma=0.6$ & $\sigma=0.65$ \\ \midrule
        HeyGen          & 87.8 & 90.1 & 91.8 & 92.1 & 91.7 \\ 
        DDIM            & 90.9 & 90.7 & 89.3 & 88.1 & 87.6 \\ 
        DiT             & 73.8 & 72.4 & 70.5 & 68.4 & 66.3 \\ 
        MidJourney-6    & 88.6 & 87.7 & 78.7 & 63.4 & 48.0 \\
        PixArt          & 83.4 & 86.3 & 87.8 & 88.7 & 89.0 \\
        SD2.1           & 96.1 & 96.8 & 96.5 & 96.1 & 95.5 \\ 
        SiT             & 75.6 & 74.5 & 72.8 & 71.1 & 68.6 \\ 
        StyleGAN2       & 90.6 & 92.5 & 93.3 & 94.0 & 94.4 \\ 
        StyleGAN3       & 92.2 & 93.4 & 93.3 & 93.1 & 92.6 \\ 
        StyleGAN-XL     & 77.3 & 75.7 & 73.2 & 71.2 & 69.9 \\
        VQGAN           & 98.5 & 99.0 & 99.0 & 99.0 & 99.0 \\
        Whichisreal     & 76.4 & 78.0 & 79.3 & 80.2 & 80.6 \\
        CollabDiff      & 98.4 & 98.4 & 97.0 & 94.9 & 92.3 \\ 
        e4e             & 99.1 & 98.7 & 97.7 & 97.4 & 97.7 \\
        StarGAN2        & 69.3 & 68.9 & 69.0 & 69.4 & 70.1 \\
        StyleCLIP       & 98.8 & 99.0 & 98.5 & 97.8 & 97.0\\ \midrule
        Avg.            & 87.3 & 87.6 & 86.7 & 85.3 & 83.8\\ \bottomrule
    \end{tabular}
    \caption{\textbf{Experiments on different parameter for Gaussian blur.}}
    \label{tab:blur_full}
\end{table*}

\begin{table*}
    \centering
    \begin{tabular}{ccccccccccc}
        \toprule
                 & \multirow{2}{*}{Dataset} & \multicolumn{3}{c}{$\sigma=0.45$} & \multicolumn{3}{c}{$\sigma=0.55$} & \multicolumn{3}{c}{$\sigma=0.65$} \\ \cmidrule(lr){3-5} \cmidrule(lr){6-8} \cmidrule(lr){9-11}
                 &  & Blur & Sharp & Contrastive & Blur & Sharp & Contrastive & Blur & Sharp & Contrastive \\ \midrule
        \multirow{16}{*}{\rotatebox{90}{DF40}}
        &HeyGen          & 87.7 & 86.0 & 88.0 & 91.8 & 87.1 & 91.7 & 91.7 & 87.6 & 90.3 \\ 
        &DDIM            & 90.9 & 90.9 & 92.3 & 89.3 & 91.7 & 92.4 & 87.6 & 91.9 & 92.0 \\ 
        &DiT             & 73.5 & 75.8 & 76.8 & 70.5 & 73.7 & 76.6 & 66.3 & 71.8 & 75.2 \\ 
        &MidJourney-6    & 88.6 & 43.2 & 81.3 & 78.7 & 38.0 & 87.0 & 47.0 & 71.8 & 75.2 \\
        &PixArt          & 83.4 & 64.8 & 78.1 & 87.8 & 59.2 & 82.3 & 89.0 & 58.6 & 84.4 \\
        &SD2.1           & 96.1 & 93.2 & 96.4 & 96.5 & 92.0 & 96.3 & 95.5 & 90.7 & 95.5 \\ 
        &SiT             & 75.6 & 78.7 & 79.0 & 72.8 & 76.5 & 78.6 & 68.6 & 74.5 & 77.1 \\ 
        &StyleGAN2       & 90.6 & 80.7 & 88.8 & 93.3 & 78.5 & 91.2 & 94.4 & 78.0 & 92.5 \\ 
        &StyleGAN3       & 92.2 & 83.7 & 91.8 & 93.3 & 78.9 & 93.0 & 92.6 & 76.8 & 92.8 \\ 
        &StyleGAN-XL     & 77.3 & 80.2 & 81.2 & 73.2 & 80.9 & 81.3 & 69.9 & 80.9 & 80.5 \\
        &VQGAN           & 98.5 & 97.7 & 99.1 & 99.0 & 97.5 & 99.4 & 99.0 & 97.0 & 99.5 \\
        &Whichisreal     & 76.4 & 79.6 & 80.3 & 79.3 & 82.1 & 83.7 & 80.6 & 83.8 & 85.1 \\
        &CollabDiff      & 98.4 & 97.4 & 99.3 & 97.0 & 97.0 & 98.3 & 92.3 & 96.0 & 96.8 \\ 
        &e4e             & 99.1 & 98.7 & 99.6 & 97.7 & 98.8 & 98.9 & 97.7 & 98.4 & 98.2 \\
        &StarGAN2        & 69.3 & 68.0 & 71.0 & 69.0 & 71.3 & 73.3 & 70.1 & 72.7 & 73.7 \\
        &StyleCLIP       & 98.8 & 96.6 & 98.8 & 98.5 & 95.6 & 98.1 & 97.0 & 94.8 & 96.8\\ \cmidrule{2-11}
        &Avg.            & 87.3 & 82.2 & 87.6 (+0.3) & 86.7 & 81.2 & 88.9 (+2.2) & 83.8 & 82.8 & 87.8 (+4.0)\\ \bottomrule \toprule
        \multirow{9}{*}{\rotatebox{90}{GenImage}}
        &ADM             & 60.1 & 51.8 & 57.1 & 57.9 & 50.3 & 53.8 & 62.3 & 50.2 & 54.3 \\
        &BigGAN          & 35.3 & 26.0 & 32.7 & 49.3 & 27.0 & 41.9 & 61.5 & 29.0 & 49.3 \\
        &Glide           & 24.6 & 25.3 & 26.3 & 30.0 & 26.3 & 30.9 & 38.4 & 27.1 & 35.8 \\
        &MidJourney      & 47.0 & 44.8 & 46.5 & 43.7 & 44.6 & 44.8 & 43.7 & 44.6 & 44.9 \\
        &SD1.4           & 62.7 & 58.7 & 61.0 & 56.6 & 58.7 & 58.2 & 54.5 & 58.9 & 57.5 \\
        &SD1.5           & 61.5 & 57.3 & 59.7 & 55.6 & 57.5 & 56.9 & 53.3 & 57.7 & 56.2 \\
        &VQDM            & 75.4 & 74.6 & 77.0 & 73.8 & 74.3 & 75.4 & 80.2 & 74.5 & 77.4 \\
        &Wukong          & 68.3 & 64.9 & 66.7 & 64.4 & 64.6 & 64.7 & 63.0 & 64.5 & 64.2 \\ \cmidrule{2-11}
        &Avg.            & 54.4 & 50.4 & 53.4 (-1.0) & 53.9 & 50.4 & 53.3 (-0.6) & 57.1 & 50.8 & 54.9 (-2.2)\\ \bottomrule
    \end{tabular}
    \caption{\textbf{Experiments on different parameters for Contrastive Blur.} The complete results of Table.~\ref{tab:contrastive}.}
    \label{tab:contrastive_full}
\end{table*}

\begin{table*}
    \centering
    \begin{tabular}{ccccccc}
        \toprule
                 \multirow{3}{*}{$\sigma_{\text{Gaussian noise}}$} & \multicolumn{6}{c}{$\sigma_{\text{Gaussian blur}}=0.45$}\\ \cmidrule(lr){2-7}
                 &  \multicolumn{3}{c}{MAX} & \multicolumn{3}{c}{MIN} \\ \cmidrule(lr){2-4} \cmidrule(lr){5-7}
                 & DF40 & GenImage & Avg. & DF40 & GenImage & Avg. \\ \midrule
                 0.001 & 71.5 & 75.8 & 73.7 & 86.2 & 55.9 & 71.1\\
                 0.002 & 70.9 & 77.1 & 74.0 & 85.0 & 63.6 & 74.3\\
                 0.004 & 79.9 & 79.1 & 79.5 & 78.7 & 74.6 & \textbf{76.7}\\
                 0.005 & 80.7 & 79.9 & 80.3 & 83.8 & 59.8 & 71.8 \\
                 0.006 & 82.6 & 79.9 & 81.3 & 82.3 & 61.9 & 72.1 \\
                 0.007 & 83.8 & 79.8 & 81.8 & 81.1 & 63.7 & 72.4 \\
                 0.008 & 84.5 & 79.3 & \underline{81.9} & 79.9 & 65.3 & 72.6 \\
                 0.009 & 85.3 & 78.6 & \textbf{82.0} & 78.7 & 66.6 & 72.7 \\ 
                 0.010 & 85.0 & 73.8 & 79.4 & 83.8 & 59.8 & 71.8 \\ 
                 0.020 & 86.2 & 61.7 & 74.0 & 72.6 & 78.2 & \underline{75.4}\\ \bottomrule

    \end{tabular}
    \caption{\textbf{Experiments on different parameters for multi-noise strategy in Table.~\ref{tab:ensemble}}.}
    \label{tab:ensemble_suppl}
\end{table*}


\begin{table*}
    \centering
    \begin{tabular}{cccccc}
        \toprule
                & Noise & Blur & MIX & MAX & MIN \\ \midrule

HeyGen          & 29.7 & 91.8 & 72.4 & 73.2 & 51.6 \\ 
DDIM            & 77.6 & 89.3 & 92.2 & 78.2 & 92.4 \\ 
DiT             & 58.3 & 70.5 & 64.9 & 58.8 & 76.6 \\ 
MidJourney-6    & 44.5 & 78.7 & 45.9 & 70.9 & 66.5 \\
PixArt          & 90.2 & 87.8 & 72.6 & 90.1 & 82.3 \\
SD2.1           & 78.5 & 96.5 & 85.9 & 83.5 & 96.0 \\ 
SiT             & 60.7 & 72.8 & 69.1 & 61.3 & 78.5 \\ 
StyleGAN2       & 88.2 & 93.3 & 97.0 & 88.1 & 91.2 \\ 
StyleGAN3       & 98.4 & 93.3 & 91.1 & 98.9 & 93.4 \\ 
StyleGAN-XL     & 81.1 & 73.2 & 78.4 & 81.1 & 81.3 \\
VQGAN           & 93.7 & 99.0 & 94.0 & 95.1 & 99.5 \\
Whichisreal     & 86.5 & 79.3 & 80.9 & 84.1 & 89.6 \\
CollabDiff      & 41.1 & 97.0 & 87.3 & 54.1 & 96.7 \\ 
e4e             & 95.5 & 97.7 & 98.9 & 95.6 & 99.0 \\
StarGAN2        & 75.6 & 69.0 & 74.2 & 75.5 & 69.0 \\
StyleCLIP       & 66.7 & 98.5 & 91.8 & 73.6 & 97.6 \\ \midrule
Avg.            & 70.1 & 86.7 & 81.0 & 78.8 & 85.3 \\  \bottomrule \toprule
ADM             & 91.4 & 57.9 & 74.1 & 67.9 & 89.1 \\
BigGAN          & 98.8 & 49.3 & 77.4 & 82.0 & 92.8 \\
Glide           & 97.8 & 30.0 & 56.2 & 75.3 & 83.6 \\
MidJourney      & 74.0 & 43.7 & 47.8 & 47.7 & 73.6 \\
SD1.4           & 64.9 & 56.6 & 58.3 & 58.1 & 65.6 \\
SD1.5           & 65.0 & 55.7 & 55.5 & 56.9 & 65.7 \\
VQDM            & 93.8 & 73.8 & 74.3 & 81.6 & 93.7 \\
Wukong          & 63.6 & 64.4 & 64.4 & 64.4 & 64.6 \\ \midrule
Avg.            & 80.3 & 53.9 & 53.9 & 66.6 & 78.6 \\ \bottomrule
    
    \end{tabular}
    \caption{\textbf{Experiments on MINDER with deepfake dataset DF40 \cite{df40} and general dataset GenImage \cite{zhu2024genimage}. The complete results of Table.\ref{tab:ensemble}.}}
    \label{tab:ensemble_full}
\end{table*}

\end{document}